\pdfoutput=1
\documentclass{article}

\usepackage[utf8]{inputenc} %
\usepackage[T1]{fontenc}    %
\usepackage[hyperfootnotes=false]{hyperref} %
\usepackage{url} %
\usepackage{booktabs} %
\usepackage{amsfonts} %
\usepackage{nicefrac} %
\usepackage{microtype} %
\usepackage{xcolor}  %
\usepackage{soul}

\newcommand{\method}[1]{\textsc{CRISP}}
\usepackage{amsmath}
\usepackage{amssymb}
\usepackage{booktabs}
\usepackage{graphicx}
\usepackage{stmaryrd}
\usepackage{multirow}
\usepackage{subfigure}
\usepackage{graphicx}
\usepackage{subcaption}
\usepackage{geometry}
\usepackage{array}

\usepackage{iclr2027_conference,times}
\iclrfinalcopy %
\hypersetup{
  pdftitle={CRISP: Compositional Relations as Invariant Structural Priors for Domain Generalization},
  pdfauthor={Dat Nguyen, Duy Nguyen},
  pdfsubject={Domain generalization, image classification, compositional spatial relations},
}

\usepackage{amsmath,amsfonts,bm}

\def\eqref#1{equation~\ref{#1}}

\def\1{\bm{1}}

\DeclareMathAlphabet{\mathsfit}{\encodingdefault}{\sfdefault}{m}{sl}
\SetMathAlphabet{\mathsfit}{bold}{\encodingdefault}{\sfdefault}{bx}{n}

\title{
CRISP: Compositional Relations as Invariant Structural Priors for Domain Generalization
}

\author{%
  Dat Nguyen\thanks{Equal contribution.} \\
  Basis Research Institute \\
  Harvard University \\
  \texttt{dat@basis.ai} \\
  \texttt{datnguyen@seas.harvard.edu} \\
  \And
  Duy Nguyen\footnotemark[1] \\
  Centre for AI Research, VinUniversity \\
  Hanoi University of Science and Technology\\
  \texttt{duy.nd@vinuni.edu.vn} \\
  \texttt{duy.nd223435@sis.hust.edu.vn} \\
}

\begin{document}

\maketitle
\lhead{Preprint}

\begin{abstract}
Domain generalization requires identifying stable representations that support reliable classification across domains. 
Domains may differ in low-level attributes, such as color, texture, or visual style, while preserving the same structural relationships among their underlying components. 
Existing methods primarily address these differences by improving the training process or aligning features across domains. However, since they leave this shared compositional structure implicit, they may overlook a more reliable source of invariance and consequently generalize less effectively to unseen domains.

We propose \textbf{C}ompositional \textbf{R}elational \textbf{I}nvariance from \textbf{S}patial \textbf{P}rimitives (\method{}), an image classification framework that factors visual recognition into visual primitives and their relational composition.
We represent these compositions using soft unary, binary, and ternary predicates over primitive locations and appearance, yielding differentiable measures of spatial and visual alignment that can be learned end-to-end. To learn primitives and relational structure jointly, we design an end-to-end architecture with three components: (1) a visual backbone that extracts generalized features, (2) a concept bottleneck layer that maps these features to primitive heatmaps with differentiable spatial coordinates, and (3) a structural scoring layer that evaluates candidate spatial relations among the detected primitives. Finally, we compute class probability from the joint evidence of its class-specific relational compositions and localized primitive appearance.

We evaluate \method{} on five real-world image-classification datasets from the widely used DomainBed suite, covering shifts in depiction style, dataset provenance, and camera-trap location. To assess its ability to exploit compositional structure, we additionally evaluate it on CUB-DG, a fine-grained bird-recognition benchmark in which distinguishing species depends on combinations of part-level attributes.
Across CUB-DG and the DomainBed benchmark suite, CRISP improves accuracy by a large margin of 7.9 percentage points on CUB-DG and 1.2 percentage points over existing DG methods on DomainBed, achieving the new state-of-the-art on both benchmarks.

\end{abstract}

\section{Introduction}\label{sec:introduction}
A central goal in image classification is to learn classifiers that remain accurate beyond the domains on which they are trained.
In practice, however, even small changes in camera, lighting, viewpoint, or style can cause a classifier’s performance to degrade~\citep{venkateswara2017deep, saenko2010adapting, nguyen2025virda, zhang2022towards}.
Such failures suggest that classifiers often exploit cues that are predictive in the training domain but unstable under distribution shift.

Domain Generalization (DG) has emerged as one of the most promising approaches to this problem, in which a model is trained on multiple labeled source datasets and evaluated on an unlabeled target dataset. Since the target domain is unavailable during learning, a DG method must determine which aspects of the image-to-label mapping are stable enough to transfer.

Relying on strong pretrained backbones~\citep{he2016deep, krizhevsky2012imagenet}, existing DG methods answer this question by modifying the training objective or optimization procedure.
Specifically, this includes aligning feature distributions across domains \citep{Sun2016DeepCORAL, Li_2018_CVPR}, perturbing or mixing samples and feature statistics \citep{zhang2018mixup, zhou2021domain, nam2021sagnet}, selecting or regularizing solutions that generalize better across domains \citep{cha2021swad, cha2022miro}, and exploiting pretrained semantic representations \citep{shu2021open}. Indeed, strong empirical studies show that even carefully tuned ERM remains competitive in standard DG benchmarks \citep{Teterwak2025ERMPlusPlus}.

Although these strategies can be effective, they largely preserve the standard recognition pipeline and rely on backbone representations to capture layout, part structure, and spatial relations implicitly. This leaves structural composition under-specified, despite longstanding evidence of its role in recognition~\citep{biederman1987recognition, felzenszwalb2009object}, and may limit performance on compositional domain generalization settings.

As a motivating example, consider images of the same bird species from two domains, Photo and Cartoon. Across these domains, local parts such as the eyes, beak, and wings may change in appearance. For example, they may be rendered with flattened colors, simplified textures, or missing specular highlights. Nevertheless, their coarse spatial organization often remains recognizable: the beak typically lies near the eyes and along the head, while the wings appear in characteristic positions relative to the body. This suggests that a classifier can benefit from distributing evidence between local primitive appearance and compositional structure, rather than forcing primitive detectors alone to absorb domain-specific variation.

In this work, we introduce \textbf{C}ompositional \textbf{R}elational \textbf{I}nvariance from \textbf{S}patial \textbf{P}rimitives (\method{}), which encourages structural invariance by representing visual categories through spatial relations among learned, localized visual primitives.
To express these relations, \method{} uses soft, differentiable spatial predicates: binary predicates capture pairwise constraints, while ternary predicates approximate layout cues such as triangular arrangements and angular constraints.
Given these relational predicates, we then represent each class as a sparse set of class-specific spatial compositions, and its score is computed by marginalizing over possible assignments of detected primitives to these compositions.
Although related predicates have been used in neuro-symbolic visual reasoning as grounded scene representations~\citep{johnson2017clevr, liu2019clevr, hsu2023s}, our goal is different: we use these predicates not as semantic reasoning targets, but as a differentiable structural inductive bias for domain-generalized classification.

This formulation creates an architectural challenge: \method{} must learn primitive detectors and classification-relevant spatial predicates jointly from image-level supervision.
We address this challenge with an end-to-end architecture that learns primitive detectors, structural predicate parameters, and class-specific compositions in a single model.
The architecture comprises three components: a visual backbone that extracts visual features, a concept bottleneck layer~\citep{koh2020concept} that maps these features to primitive heatmaps with differentiable spatial coordinates, and a structural scoring layer that evaluates candidate spatial relations among the detected primitives. The structural scoring layer encodes the space of possible relational compositions in a differentiable manner, enabling joint optimization of primitive detectors and structural weights via standard gradient descent. The resulting relation scores are aggregated by a final classifier, producing predictions grounded in interpretable compositions of visual primitives and their spatial arrangements.

We evaluate \method{} on the widely used DomainBed benchmark~\citep{gulrajanisearch} and CUB-DG~\citep{Min2022GVRT}, a benchmark for compositional domain generalization. On DomainBed, \method{} achieves accuracies of 70.1\% and 79.0\% with ResNet50 and ViT-B backbones, respectively, surpassing the state-of-the-art method ERM++~\citep{Teterwak2025ERMPlusPlus} by 1.2 and 0.4 percentage points. On CUB-DG, \method{} yields a larger improvement of 7.9 percentage points with ResNet50, highlighting its effectiveness in targeting compositional domain shift.

We summarize our contributions as follows:
\begin{itemize}
\item We propose a structure-aware framework for domain generalization that treats visual categories as compositions of learned primitives and spatial relations, improving performance on compositional recognition under domain shift.
\item We introduce an end-to-end differentiable architecture that realizes this framework by learning primitive heatmaps, extracting differentiable spatial coordinates, and scoring class-specific compositions over primitive locations.
\item We evaluate \method{} on the DomainBed and CUB-DG benchmark suite, showing that it improves over the prior state of the art on both benchmark suites by 1.2\% and 7.9\%, respectively, using the traditional ResNet50 backbone while remaining competitive on the modern ViT-B backbone; our ablation studies further show that the proposed structural scoring layer consistently improves performance across domains and datasets.
\end{itemize}

The remainder of the paper is organized as follows.
Section~\ref{sec:related_works} reviews related work on domain generalization, spatial predicates, and concept learning.
Section~\ref{sec:method} presents \method{}.
Section~\ref{sec:experiments} reports the experimental setup and results.
Section~\ref{sec:conclusion} concludes with limitations and future directions.

\section{Related Work}\label{sec:related_works}

This section reviews three lines of work most relevant to \method{}: domain generalization, spatial relations and predicates for visual recognition, and concept bottleneck learning.

\paragraph{Domain generalization and adaptation.}
Domain adaptation and domain generalization both address distribution shift, but differ in target-domain access. Unsupervised domain adaptation assumes unlabeled target samples are available during training, and commonly reduces source-target discrepancy through distribution matching or adversarial objectives~\citep{long2015learning, Sun2016DeepCORAL, Ganin2016DANN, Long2018CDAN}. Recent extensions further improve adaptation with transformer architectures, parameter-efficient modules, or vision-language prompting~\citep{sun2022safe, xu2022cdtrans, nguyen2025virda, bai2024prompt, Khattak_2023_CVPR}.

Domain generalization is stricter: target domains are unseen during training. Existing DG methods seek transferable representations or training procedures through invariant feature learning, risk or optimization-based regularization, and style or feature perturbation~\citep{muandet2013domain, arjovsky2019invariant, krueger2021out, cha2021swad, cha2022miro, zhou2021domain, nam2021sagnet}. While carefully tuned ERM remains a strong baseline on standard DG benchmarks~\citep{gulrajanisearch}, CUB-DG highlights compositional shifts in both appearance and visual structure~\citep{Min2022GVRT}. \method{} complements these feature-centric approaches by explicitly scoring compositions of learned primitives and differentiable spatial predicates.

\paragraph{Spatial relations and predicates for visual recognition.}
Spatial structure has long been central to recognition, from classical part-relation theories~\citep{biederman1987recognition} and deformable part models~\citep{felzenszwalb2009object} to neural relation modules over objects or regions~\citep{Hu_2018_CVPR}. Predicates provide a compact way to express such structure and are widely used in neuro-symbolic visual reasoning for relations such as position, proximity, and containment~\citep{johnson2017clevr, mao2019neuro}. Recently, NeSyCoCo studies differentiable composition of predicate scores for compositional generalization in vision-language reasoning~\citep{Kamali_Barezi_Kordjamshidi_2025}. In contrast to these works, \method{} uses differentiable spatial predicates not as grounded semantic reasoning targets or language-conditioned programs, but as an inductive bias for domain-generalized classification, providing a vocabulary for scoring spatial compositions among learned visual primitives.

\paragraph{Concept bottleneck learning.}
Concept bottleneck models predict through intermediate concept representations, often using predefined human-interpretable concepts with concept-level supervision~\citep{koh2020concept}. Another thread studies concept discovery and concept-based explanations under weaker supervision~\citep{chen2020concept}. \method{} adopts the bottleneck idea for spatial primitive learning: it learns localized primitive heatmaps from image-level supervision and derives differentiable spatial descriptors from them. Unlike standard concept bottlenecks that classify from attribute-like concept vectors, \method{} uses these descriptors as localized evidence for differentiable relational compositions in the structural scoring layer.

\section{Methodology}\label{sec:method}
\begin{figure}
    \centering
    \includegraphics[width=1\textwidth]{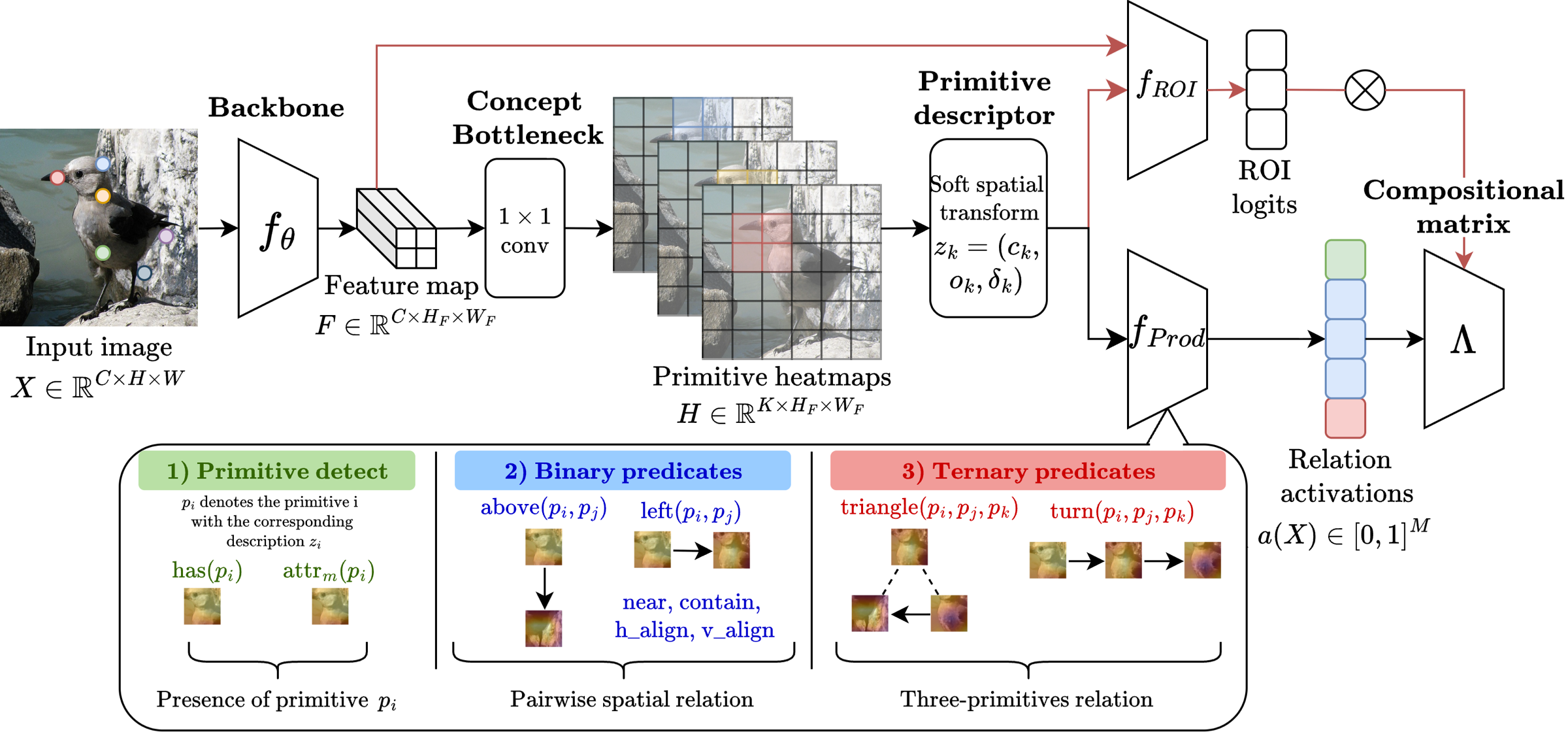}
    \caption{The overall pipeline of \method{}.}
    \label{fig:model}
\end{figure}

In this section, we describe the methodology of \method{}.
We consider the standard domain generalization setting, in which we are given a training set $\mathcal{D} = \{(X_i, Y_i, d_i)\}_{i=1}^N$, where $X_i \in \mathbb{R}^{3 \times H \times W}$ is an image, $Y_i \in \mathcal{C}$ is its class label, and $d_i \in \mathcal{S}$ indicates its source domain.
At test time, the model is evaluated on samples from an unseen target domain $t \not\in \mathcal{S}$ that shares the same label space $\mathcal{C}$ as the source domains but was not observed during training.

As in prior DG work, our goal is to train a classifier that generalizes to unseen target domains. Furthermore, 
instead of relying solely on backbone representations, we explicitly model visual primitives and their spatial relations, while learning both end-to-end from image-level supervision.
We first describe the primitive-and-structure component of \method{} in Section~\ref{sec:structure}, where we formally define visual primitives and the spatial predicates over their locations.
Then, we describe our network architecture that allows jointly learning both these structures and primitives in Section~\ref{sec:architecture}. 
In the same section, we also describe how primitive heatmaps are extracted from backbone features, how spatial predicates 
are computed over primitive locations in this network, and how we combine both spatial predicates and the backbone features for classification.

\subsection{Representing Primitives and Structure}\label{sec:structure}

\paragraph{Visual primitives.} To represent visual primitives, we assume that there exists a set of $K$ learned visual primitives $\mathcal{P}=\{p_1,\ldots,p_K\}$.
Each primitive $p_k: X \mapsto z_k(X)$ is a mapping from any input image $X$ to the corresponding image-dependent descriptor $z_k(X)=\langle c_k(X), \sigma_k(X), \delta_k(X), \ell_k(X)\bigr\rangle$, with associated spatial location $c_k(X)\in\mathbb{R}^2$, a presence score $\sigma_k(X)\in[0,1]$,  the corresponding spatial extent $\delta_k(X)\in\mathbb{R}^2$, and finally, a vector for appearance descriptor $\ell_k(X)$. The primitives should be learned from image-level supervision and should not require manual annotations, which we leave for Section~\ref{sec:architecture} to describe how $p_k$ and $z_k(X)$ are parameterized and optimized in the full network.
Below, we define spatial predicates over these image-dependent primitive descriptors, which are used to form class-specific structural compositions.

\paragraph{Differentiable spatial predicates.}
Given the set of primitives $\mathcal{P}$, we model their spatial relations as a collection of soft spatial predicates $\mathcal{R}$ with varying arity. Each predicate would takes as input a tuple of primitives $\langle p_1, \ldots, p_r \rangle$ (where $r$ is the arity of the predicate) along with an input image $X$ and output a soft satisfaction score in the range of $[0, 1]$: 
$$
R: \langle p_1, \ldots, p_r \rangle \times X \mapsto \text{score}_R \in [0, 1]
$$
Implementation-wise, all predicates operate over the image-dependent description $z_k(X)$ for each primitive $p_k$. Thus, for simplicity, in the rest of this section, we omit the dependence on $X$: We use $z_k = \langle c_k, \sigma_k, \delta_k, \ell_k \rangle$ in place of $z_k(X)$. Furthermore, we expand $c_k = \langle c_k^x, c_k^y \rangle$ and $\delta_k = \langle \delta_k^x, \delta_k^y \rangle$, respectively.
For predicates that require box corners, we derive a soft bounding box $b_k$ from the primitive location $c_k$ and extent $\delta_k$: 
\[
b_k =
\langle 
b_k^{x_1}, b_k^{y_1}, b_k^{x_2}, b_k^{y_2}
\rangle
=
\langle
c_k^x-\delta_k^x,\;
c_k^y-\delta_k^y,\;
c_k^x+\delta_k^x,\;
c_k^y+\delta_k^y
\rangle.
\]
Finally, before defining the predicates, we make two remarks about the parameterization of these predicates. Firstly, all predicate parameters we describe below, including $\kappa_{(\ldots)}$, $m_{(\ldots)}$, $\tau_{(\ldots)},\rho, \beta,\eta,\gamma, \theta$ and $\phi$ are all learnable. Secondly, we parameterize each predicate by its family; for example, the above predicate $R_{\text{above}}$'s parameters will be shared across different combinations of input primitives, and thus, the set of learnable parameters for all these predicates is fixed with respect to the number of primitives and the number of classes.
We now describe each spatial predicate in detail below. 

\paragraph{Primitive presence.}

Spatial relations describe how primitives are arranged, while appearance cues help distinguish primitives with similar spatial configurations. Since patterns such as curved or dashed contours can recur across different primitives, we introduce a shared vocabulary of $N_\text{attr}$ learned unary appearance predicates. This vocabulary allows the model to reuse appearance tests across primitives and combine them with different spatial relations. Implementation-wise, for each primitive $p_k$ and its corresponding appearance descriptor $\ell_k \in \mathbb{R}^d$, we apply $N_{attr}$ scalar affine heads to this descriptor:
$$
R_{\mathrm{attr}}^{(n)}(p_k)
=
\mathrm{sigmoid}\!\left(
w_n^{\top}\mathrm{LayerNorm}(\ell_k)+b_n
\right),
\qquad n\in\{1,\ldots,N_\text{attr}\},
$$
where \(w_n\in\mathbb{R}^{d}\) and \(b_n\in\mathbb{R}\) are learned parameters. Each vocabulary element $m$ thus applies the same learned appearance test to every primitive, producing a soft predicate activation
$R_{\mathrm{attr}}^{(n)}(p_k) \in [0, 1]$. A primitive can activate multiple predicates, allowing its appearance to be represented through combinations of these shared vocabulary elements. The structural layer, as we would describe later, combines these activations with spatial predicates to form class-specific structural compositions.

\paragraph{Binary predicates.} We use binary predicates to encode pairwise spatial relations between two primitives $p_i$ and $p_j$.
They correspond to common spatial relations used in predicate-based visual reasoning, such as relative position, alignment, proximity, and containment~\cite{johnson2017clevr, yi2018neural, mao2019neuro}. We give their detailed formula below:
\[
\begin{array}{ll}
R_{\mathrm{above}}(p_i,p_j)
= \operatorname{sigmoid}\!\left(\kappa_{\uparrow}(c_j^y-c_i^y-m_{\uparrow})\right)
&
R_{\mathrm{left}}(p_i,p_j)
= \operatorname{sigmoid}\!\left(\kappa_{\leftarrow}(c_j^x-c_i^x-m_{\leftarrow})\right)
\\[1.0em]

R_{\mathrm{h\text{-}align}}(p_i,p_j)
= \exp\!\left(-\dfrac{(c_i^y-c_j^y)^2}{2\tau_h^2}\right)
&
R_{\mathrm{v\text{-}align}}(p_i,p_j)
= \exp\!\left(-\dfrac{(c_i^x-c_j^x)^2}{2\tau_v^2}\right)
\\[1.2em]

\multicolumn{2}{l}{
R_{\mathrm{near}}(p_i,p_j)
= \exp\!\left(
-\dfrac{(c_i^x-c_j^x)^2+(c_i^y-c_j^y)^2}{2\rho^2}
\right)
}
\\[1.2em]

\multicolumn{2}{l}{
R_{\mathrm{contains}}(p_i,p_j)
= \operatorname{sigmoid}\!\left(
\kappa_{\supset}
\min\!\left[
b_j^{x_1}-b_i^{x_1},\,
b_j^{y_1}-b_i^{y_1},\,
b_i^{x_2}-b_j^{x_2},\,
b_i^{y_2}-b_j^{y_2}
\right]
\right).
}
\end{array}
\]
Here, $R_{\mathrm{above}}$ and $R_{\mathrm{left}}$ score directional relations, indicating whether $p_i$ is above or to the left of $p_j$ in the image, respectively.
The alignment predicates, $R_\mathrm{h-align}$ and $R_\mathrm{v-align}$ score whether two primitives share approximately the same vertical or horizontal position, while $R_{\mathrm{near}}$ scores spatial proximity. 
The containment predicate scores whether the soft bounding box of $p_i$ encloses that of $p_j$. 
The learnable parameters $\kappa$ control predicate sharpness, $m$ denotes directional margins, and $\tau_h,\tau_v,\rho$ control tolerance for alignment and proximity. 

\noindent While binary predicates capture local spatial relations, object and part configurations often depend on geometric patterns involving multiple parts, such as angles, triangular arrangements, and relative distances between relations themselves (i.e., higher-order relations)~\citep{biederman1987recognition, felzenszwalb2009object}.
While it is desirable to model these relations, modeling higher-order relations and predicates exactly is difficult to learn and is also difficult to optimize directly. Thus, we express such patterns through enumerable ternary predicates. This keeps the structural layer differentiable and tractable, while allowing it to approximate higher-order layout cues over triples of primitives and pairs of primitive pairs. We describe this in detail below.
\paragraph{Ternary predicates.} We model three-primitive geometric cues with two ternary predicates, $R_\text{tri}$ and $R_\text{turn}$. 
For primitives $p_i,p_j,p_k$, let $\alpha_{ijk}$ denote the interior angle at $p_i$, and let
$\theta^{\mathrm{turn}}_{ijk}=\arccos(\hat{\mathbf{v}}_{ij}\cdot\hat{\mathbf{v}}_{jk})$
denote the turning angle along the ordered chain $p_i\!\to\!p_j\!\to\!p_k$.
We define
\[
\begin{array}{ll}
R_{\mathrm{tri}}^{(n)}(p_i,p_j,p_k)
=
\exp\!\left(
-\dfrac{(\alpha_{ijk}-\psi_n)^2}{2\beta_n^2}
\right)
&
R_{\mathrm{turn}}^{(\ell)}(p_i,p_j,p_k)
=
\exp\!\left(
-\dfrac{(\theta^{\mathrm{turn}}_{ijk}-\phi_\ell)^2}{2\eta_\ell^2}
\right).
\end{array}
\]
Here, $R_{\mathrm{tri}}^{(n)}$ scores triangular configurations with target angle $\psi_n$, while $R_{\mathrm{turn}}^{(\ell)}$ scores ordered chain turns with learnable target angle $\phi_\ell$; $\beta_n$ and $\eta_\ell$ control the corresponding tolerances.

\paragraph{Class-specific structural compositions.}
Together, these predicates define a differentiable vocabulary for spatial structure.
While predicate parameters are shared globally, each class learns a sparse weighting over the enumerated predicate--primitive assignments.
Given an image, \method{} evaluates these assignments on the detected primitives and aggregates their weighted scores into class evidence.
Thus, classification depends jointly on primitive presence and on whether the detected primitives satisfy class-specific spatial patterns.

\subsection{Network Architecture}\label{sec:architecture}

In this section, we describe the network architecture that allows modeling and learning both the aforementioned primitives and structures. We give an overview of this architecture in Figure~\ref{fig:model}.
This architecture consists of three main components: a visual backbone that allows extracting general visual features, 
a concept bottleneck layer that maps these features to the corresponding primitive along with their heatmaps and differential spatial coordinates as required by Section~\ref{sec:structure}.
Since the unary appearance predicate also requires the corresponding primitive features, we describe below how we use the region-of-interest (ROI) extractor to get these corresponding features.
Having all the predicates and primitive features, we show how our structural scoring layer combines both these two sources of information to make predictions.
We describe each of these components in turn below.

\paragraph{Visual backbone.}
Given an image 
$X\in\mathbb{R}^{3\times H\times W}$, we first use a visual backbone $f_{\theta}$ to extract a feature map
$F=f_{\theta}(X)\in\mathbb{R}^{C\times H_F\times W_F}$,
where $C,H_F,W_F$ denote the channel, height, and width dimensions of the feature map.

\paragraph{Primitive concept bottleneck.}
This module instantiates the primitive descriptors defined in Section~\ref{sec:structure}.
Inspired by concept bottleneck models~\citep{koh2020concept} and differentiable heatmap-based localization~\citep{nibali2018numerical}, we project the feature map $F$ into $K$ primitive heatmaps,
\[
F_H= g_{\mathrm{cb}}(F) \in \mathbb{R}^{K\times H_F\times W_F},
\]
where $g_{\mathrm{cb}}$ is a lightweight convolutional projection.
For image $X$, the heatmap $F_{H}^{(k)}$ is the $k^{\text{th}}$ channel of $F_H$ that provide spatial evidence for primitive $p_k$ and is converted into the descriptor
$z_k(X)=\langle c_k(X),\sigma_k(X),\delta_k(X), \ell_k(X)\rangle$.
Following differentiable heatmap-to-coordinate transforms~\citep{nibali2018numerical}, we form a temperature-normalized heatmap
\[
\widetilde{F}^{(k)}_{H}[h, w]
=
\frac{\exp(F_H^{(k)}[h, w]/T)}
{\sum_{h',w'}\exp(F_H^{(k)}[h',w']/T)},
\quad
c_k(X)
=
\sum_{h,w}\widetilde{F}_H^{(k)} \odot\ G_\text{coord},
\quad
\ell_k(X) = \sum_{h, w} \widetilde F_H^{(k)} \odot F
\]
where $T$ is a learnable temperature, $\odot$ is element-wise multiplication and
$G_\text{coord} = \left\langle
\frac{2w}{W_F-1}-1,
\frac{2h}{H_F-1}-1
\right\rangle
\in[-1,1]^2$ is the normalized grid coordinate map. We compute the presence confidence as
$\sigma_k(X)=\operatorname{sigmoid}(\max F_H^{(k)})$, and define $\delta_k(X)$ as twice the coordinate-wise standard deviation under $\widetilde{H}_{k}$.
Intuitively, \(\ell_k(X)\) is an attention-pooled appearance descriptor that aggregates backbone features from the regions emphasized by the primitive's heatmap.
These descriptors provide the locations, confidences, and extents used by the structural scoring layer.

\paragraph{Localized scoring layer.}
While spatial predicates provide a general mechanism for composing visual primitives, class identity can additionally depend on local appearance. We therefore complement the structural scoring layer with a parallel appearance voting head that aggregates direct class evidence from individual primitive regions. 
For each primitive \(p_k\), a shared detection-style ROI head \(f_{\mathrm{ROI}}\)~\citep{he2017mask} pools backbone features \(F\) within its predicted bounding box \(b_k(X)\) to a fixed \(3\times3\) grid, averages them spatially, and maps the resulting vector to class logits. We aggregate these regional votes using normalized presence confidences:
$$
\ell_{\mathrm{ROI}}(X)
=
\sum_{k=1}^{K}
\frac{\sigma_k(X)}
{\sum_{j=1}^{K}\sigma_j(X)}
\,f_{\mathrm{ROI}}\!\left(F,b_k(X)\right)
\in\mathbb{R}^{|\mathcal{C}|}.
$$
Both branches operate on the same primitive set: the appearance head produces regional class votes from features pooled within predicted boxes, while the structural layer composes spatial predicates with appearance predicates derived from heatmap-pooled features.

\paragraph{Structural scoring layer.}
The structural scoring layer converts the structural vocabulary from Section~\ref{sec:structure} into class-level evidence.
Given primitive descriptors $\{z_k(X)\}_{k=1}^K$, it evaluates unary presence scores and enumerates spatial predicate applications over valid primitive assignments:
\[
a_{R,\mathbf{i}}(X)=R(p_{i_1},\ldots,p_{i_r};X),
\quad
R\in\mathcal{R}_r,\;
\mathbf{i}=\langle i_1,\ldots,i_r\rangle\in\mathcal{A}_r.
\]
where \(\mathcal{R}_r\) denotes the set of \(r\)-ary predicates and \(\mathcal{A}_r\subseteq\{1,\ldots,K\}^{r}\) contains the valid ordered assignments of primitives to their arguments.
Collecting these activations yields \(a(X)\in[0,1]^M\), where \(M\) is the total number of predicate applications. Ternary predicates account for most applications in our configuration, with an enumeration cost of \(O(|\mathcal{R}_3|K^3)\). With \(K=16\), the layer evaluates approximately \(8{,}000\) predicate applications per image.
We parameterize class-specific structural compositions using
$
\Lambda = [\lambda^1;\lambda^2;\ldots;\lambda^{|\mathcal{C}|}]
\in \mathbb{R}^{|\mathcal{C}|\times M},
$
where \(\lambda^c\in\mathbb{R}^{M}\) assigns an unnormalized weight to each predicate activation for class \(c\). 
Unlike the globally shared predicate-shape parameters, these structural weights are class-specific.
We normalize them with sparsemax~\citep{martins2016softmax} and compute the class-wise structural score $\ell_\text{struct}(X) \in [0, 1]^\mathcal{C}$ as
\[
w^c = \operatorname{sparsemax}(\lambda^c) \in \Delta^{M-1},
\qquad
\ell_\text{struct}(X)[c] = (w^c)^\top a(X) \in [-1, 1]
\]
Where $\Delta^{M-1}$ denotes the probability simplex. Thus, each class selects a sparse subset of primitive-presence and spatial-predicate activations as its structural composition. Because these activations are differentiable, gradients from the classification loss propagate through the structural scoring layer to the primitive bottleneck and visual backbone. 

\noindent\textbf{Class scoring and learning objective}
We compute the final class scores by summing the structural and appearance scores with log-softmax for numerical stability,
$
s(X)
=
\text{logsoftmax}(\ell_{\mathrm{struct}}(X)
+
\ell_{\mathrm{ROI}}(X)
) \in\mathbb{R}^{|\mathcal{C}|}
$
and train \method{} end-to-end with:
\begin{equation}
\mathcal{L}
=
\mathcal{L}_{\mathrm{CE}}(s(X),y)
+ \mathcal{L}_{\mathrm{reg}}
+ \mathcal{L}_{\mathrm{DANN}},
\label{eq:learning_objective}    
\end{equation}
where $\mathcal{L}_{\mathrm{CE}}$ is cross-entropy, $\mathcal{L}_{\text{reg}}$ stabilizes predicate learning and structural selection, and $\mathcal{L}_\text{DANN}$ helps to make the backbone features produced by $f_{\theta}$ domain-invariant, following~\citep{Ganin2016DANN}. As the architecture is fully differentiable, this objective jointly optimizes all the aforementioned components. 
We defines $\mathcal{L}_{\mathrm{reg}}$ and $\mathcal{L}_\text{DANN}$ in details in Appendix~\ref{app:regularizers}.

\section{Experiments}\label{sec:experiments}
We employ the widely used DomainBed~\citep{gulrajanisearch, cha2021swad} protocols and the challenging new CUB-DG dataset~\citep{Min2022GVRT} for evaluating our method. The rest of the dataset description and the implementation details can be found in our Appendix~\ref{app:datasets}. In the next sections, we describe our experimental results.

\subsection{Overall Results}

\begin{table*}[t]
\centering
\small
\caption{Accuracy on DomainBed and CUB-DG. Best in \textbf{bold}; second-best in \underline{underlined}.}
\label{tab:main_results}
\vspace{-0.5em}
\setlength{\tabcolsep}{4.0pt}
\begin{tabular}{@{}clccccccc@{}}
\toprule
 & Method & PACS & VLCS & OfficeHome & TerraInc & DomainNet & CUB-DG & DomainBed \\
\midrule
\multirow{12}{*}{\rotatebox[origin=c]{90}{\textbf{ResNet-50}}}
 & ERM & 84.2 & 77.3 & 67.6 & 47.8 & 44.0 & - & 64.2 \\
 & DANN & 83.6 & 78.6 & 65.9 & 46.7 & 38.3 & 49.5 & 62.6 \\
 & CDANN & 82.6 & 77.5 & 65.8 & 45.8 & 38.3 & 48.6 & 62.0 \\
 & GroupDRO & 84.4 & 76.7 & 66.0 & 43.2 & 33.3 & 44.8 & 60.7 \\
 & MixUp & 85.2 & 77.9 & 60.4 & 44.0 & 34.0 & 50.3 & 60.3 \\
 & SagNet & 86.3 & 77.8 & 68.1 & 48.6 & 40.3 & 51.6 & 64.2 \\
 & CORAL & 86.2 & 78.8 & 68.7 & 47.7 & 41.5 & 55.5 & 64.6 \\
 & GVRT & 85.1 & 79.0 & 70.1 & 48.0 & 44.1 & 57.0 & 65.3 \\
 & MIRO & 85.4 & 79.0 & 70.5 & 50.4 & 44.3 & 49.1 & 65.9 \\
 & SWAD & 88.1 & \underline{79.1} & 70.6 & 50.0 & 46.5 & 52.2 & 66.9 \\
 & ERM++ & \textbf{89.8} & 78.0 & \textbf{74.7} & 51.2 & \textbf{50.8} & \underline{61.1} & \underline{68.9} \\
\cmidrule(lr){2-9}
 & \method{} (Ours) & \underline{88.4} & \textbf{80.4} & \underline{73.4} & \textbf{58.5} & \underline{50.0} & \textbf{69.0} & \textbf{70.1} \\
\midrule
\multirow{6}{*}{\rotatebox[origin=c]{90}{\textbf{ViT-B/14}}}
 & ERM & 83.4 & 75.9 & 66.4 & 35.3 & 44.4 & - & 61.1 \\
 & MIRO & 95.6 & 82.5 & 82.5 & 54.3 & 54.0 & - & 73.7 \\
 & CAR-FT & \underline{96.8} & \underline{85.5} & 85.7 & 61.9 & \textbf{62.5} & - & 78.5 \\
 & UNI-DG & 96.7 & \textbf{86.3} & 86.2 & 62.4 & \underline{61.3} & - & \underline{78.6} \\
 & ERM++ & \underline{96.8} & 82.4 & \underline{86.7} & \textbf{67.4} & 59.8 & \textbf{81.2} & \underline{78.6} \\
\cmidrule(lr){2-9}
 & \method{} (Ours) & \textbf{97.0} & 84.4 & \textbf{88.0} & \underline{67.0} & 58.5 & \underline{80.1} & \textbf{79.0} \\
\bottomrule
\end{tabular}
\end{table*}

We report the mean accuracy of each of the 5 datasets in the DomainBed setting, and the CUB-DG dataset in Table~\ref{tab:main_results}. As can be seen in the table, our method is compatible with different backbone types such as the traditional ResNet-50 and the modern ViT-B/14 backbone in the DomainBed setting. Specifically, \method{} surpasses the current SOTA method ERM++ using both of the backbones with 1.2\% and 0.4\% on the DomainBed setting. On CUB-DG, \method{} achieves the best mean accuracy of 69.0\%, outperforming the strongest prior result, ERM++, by 7.9 points. These results suggest that explicitly modeling spatial compositions is particularly effective on CUB-DG, where fine-grained bird categories share coarse part layouts while appearance varies substantially across domains.

\subsection{Ablation Studies}
We pose the following research questions (RQ)s as ablation studies for \method{}:
\begin{figure}
    \centering
    \includegraphics[width=0.92\linewidth]{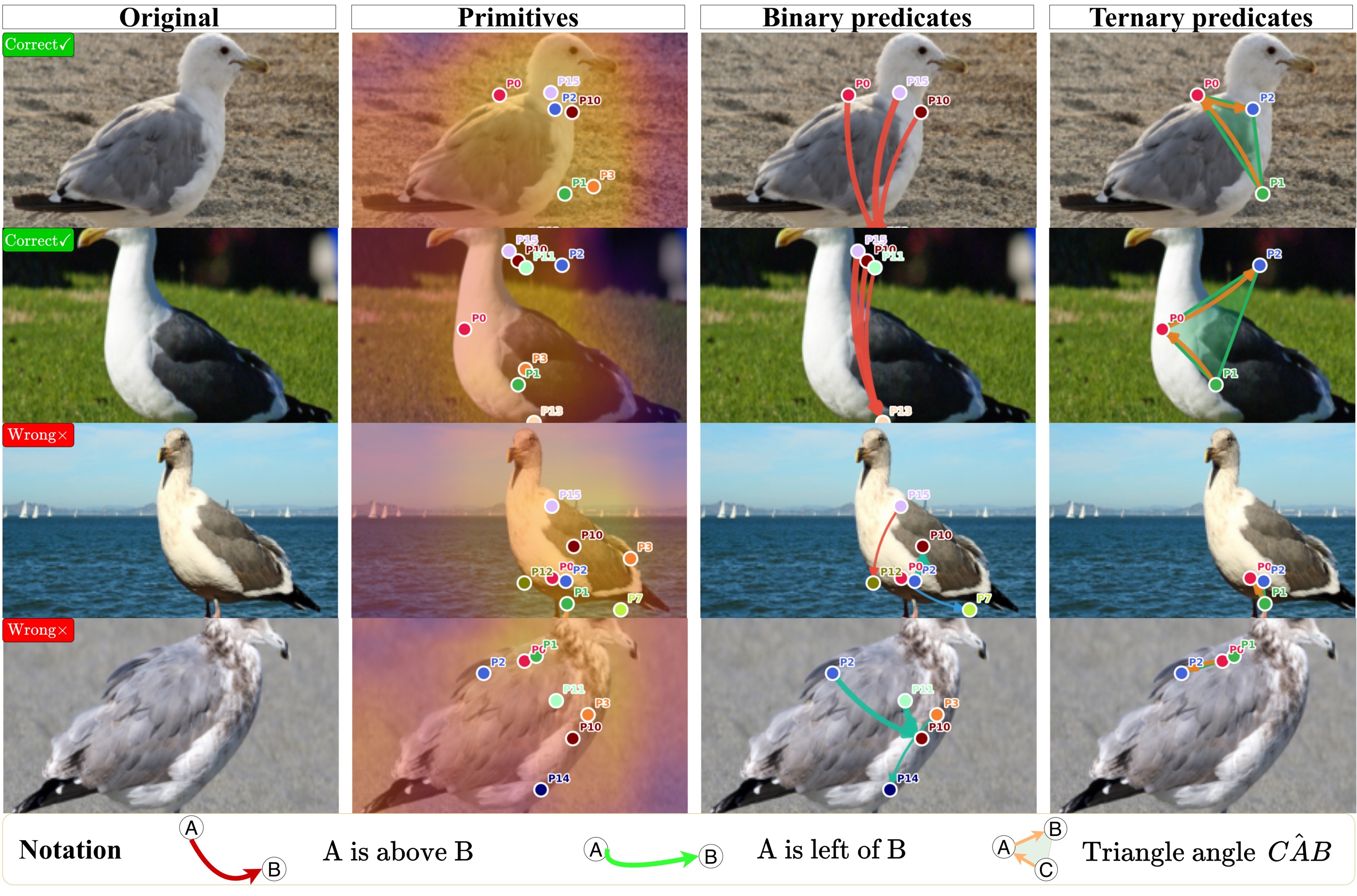}
    \caption{Error analysis on 2 correct and 2 false classification samples of the target (Photo) domain. Error happens due to the misidentification of the crucial primitive region $P_{13}$.}
    \label{fig:error}
    \vspace{-0.8em}
\end{figure}

\begin{figure}
    \centering
    \begin{minipage}[t]{0.49\textwidth}
        \centering
        \includegraphics[width=\linewidth]{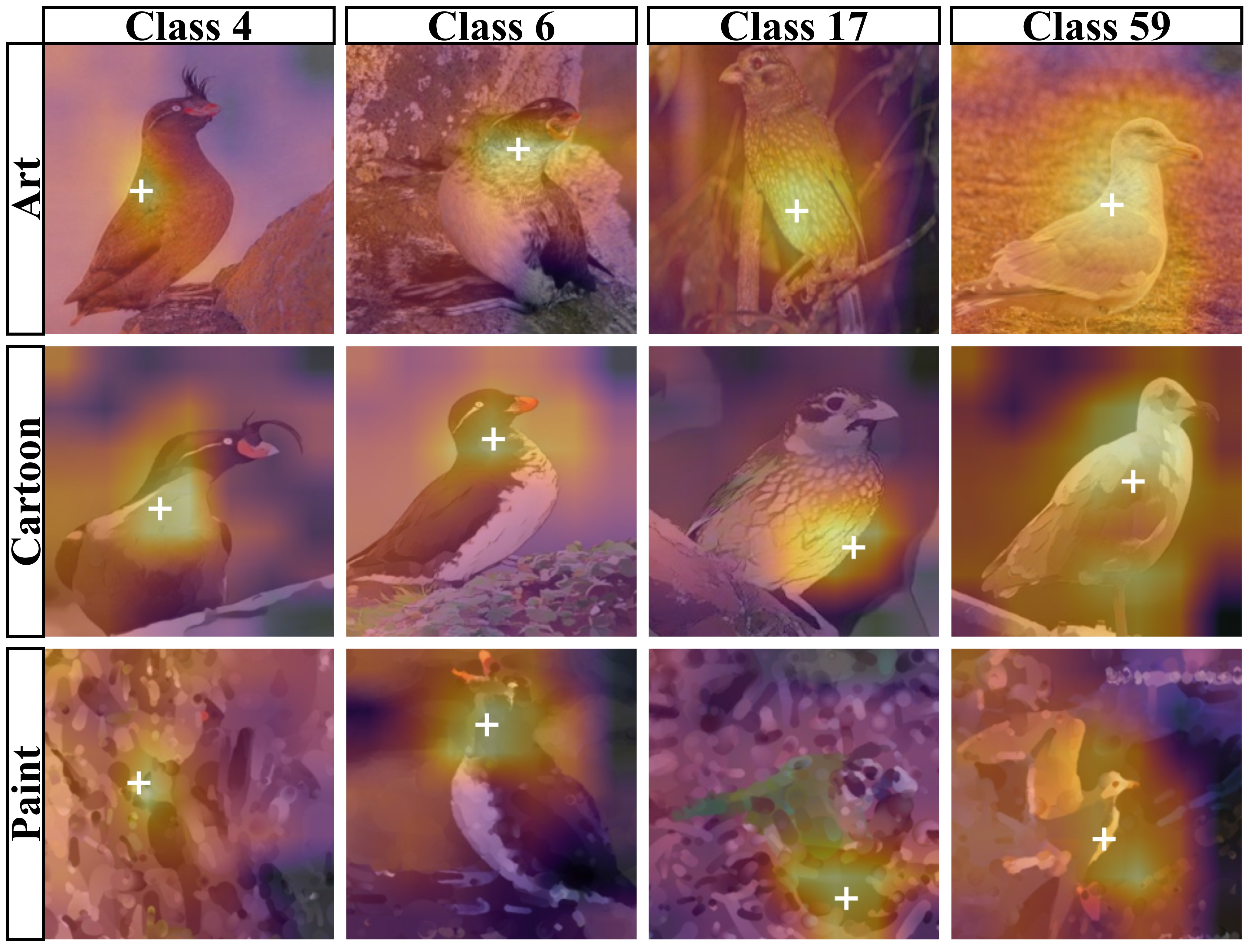} 
        \caption{Primitive 0 visualization over different classes of source domains.}
        \label{fig:primitive}
    \end{minipage}\hfill
    \begin{minipage}[t]{0.49\textwidth}
        \centering
        \includegraphics[width=\linewidth]{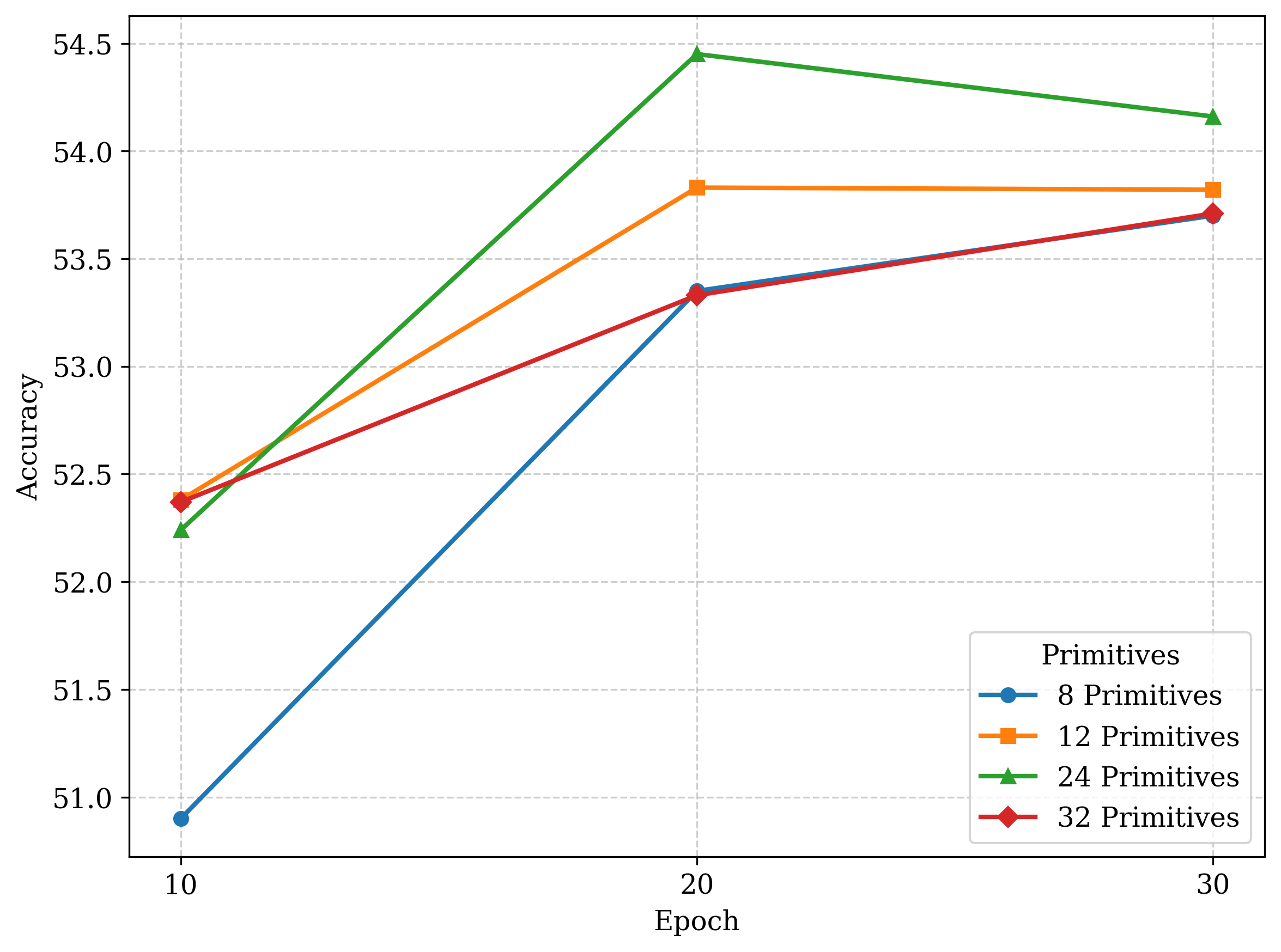} 
        \caption{Results of \method{} with different numbers of primitives on DomainNet/Sketch dataset.}
        \label{fig:ablation_prim_k}
    \end{minipage}
\end{figure}

\noindent\textbf{RQ1: What do the learned primitives capture?}
As illustrated in Fig.~\ref{fig:primitive}, \method{}'s learned primitive 0 localizes class-relevant bird regions across source domains, such as the \textit{wing-chest} region for class 4, the \textit{neck} region for class 6, and the \textit{wing-only} region for class 17. 
These localized primitives provide the building blocks for \method{} to model class-specific spatial relations. Furthermore, their localization patterns remain relatively consistent across changes in texture and pose, suggesting that the learned primitives capture spatially meaningful evidence that is robust to domain shift. We will display more closely the per-dataset results and learned primitives with their corresponding relations in Appendix~\ref{app:per-domain}.

\noindent\textbf{RQ2: What failure modes does \method{} exhibit?}
Fig.~\ref{fig:error} shows that primitive localization is a key bottleneck for \method{}. In correctly classified samples, primitives localize informative regions: for example, $P_{15}$ and $P_{10}$ focus on the neck--head region, while $P_{13}$ captures the feet. In misclassified samples, several primitives collapse around the torso and fail to anchor to class-discriminative extremities, including the region typically captured by $P_{13}$. This weakens the geometric evidence available to the structural scoring layer, making the final prediction more error-prone.

\noindent\textbf{RQ3: How does the number of primitives affect performance?}
We study the effect of the number of primitives on the DomainNet task with Sketch as the target domain, as shown in Fig.~\ref{fig:ablation_prim_k}. Increasing the number of primitives from the default setting to $K=24$ improves accuracy by 0.5 points, suggesting that additional primitives can provide finer spatial evidence. However, since the number of enumerated predicate applications grows with $K$, larger primitive sets increase the computational cost of the structural scoring layer. We therefore use the smaller default configuration for the DomainBed experiments as a trade-off between accuracy and efficiency.

\noindent\textbf{RQ4: What is the contribution of each components in \method{}?}
We dedicate Table~\ref{tab:backbone_ablation} and Table~\ref{tab:grammar_ablation_and_compaction} to study the contributions of different training strategies and network components. Specifically, Table~\ref{tab:backbone_ablation} studies whether applying ERM++~\cite{Teterwak2025ERMPlusPlus} (\method{} (ERM++)) would yield better performance. We also study the contribution of a localized scoring layer (through \method{} (w/o ROI)) and that of the structural scoring layer (through \method{} w/o ROI)), respectively. The results show that the full \method{} architecture, with both localized and structural scoring layers, outperforms its ablated baselines consistently, and on both convolutional and transformer-based backbones. Interestingly, \method{} (ERM++) gives the new state-of-the-art on the PACS dataset with the RN50 backbone. We also perform more fine-grained ablations on the predicate sets on CUB-DG in Table~\ref{tab:grammar_ablation_and_compaction} and show that adding binary, ternary, their combination, and their combinations with unary appearance predicates consistently improve \method{}'s overall performance.

\begin{table*}[t]
\centering
\begin{minipage}[t]{0.53\textwidth}
    \centering
    \caption{Contributions of network components.}
    \label{tab:backbone_ablation}
    \vspace{-0.7em}
    \footnotesize
    \setlength{\tabcolsep}{2.5pt}
    \renewcommand{\arraystretch}{1.05}
    \resizebox{\linewidth}{!}{%
    \begin{tabular}{@{}clcccccc@{}}
        \toprule
         & Method & PACS & VLCS & OH & Terra & DMN & Avg. \\
        \midrule
        \multirow{4}{*}{\rotatebox[origin=c]{90}{RN50}}
        & CRISP (ERM++)
        & \textbf{89.8} & \underline{78.4} & \underline{73.1}
        & \underline{57.8} & \underline{48.8} & \underline{69.6} \\
        & CRISP (w/o ROI)
        & 86.7 & 76.9 & 70.5 & 54.8 & 44.4 & 66.7 \\
        & CRISP (w/o Grammar)
        & 87.2 & 75.0 & 71.3 & 52.6 & 45.6 & 66.3 \\
        & CRISP (full)
        & \underline{88.4} & \textbf{80.4} & \textbf{73.4}
        & \textbf{58.5} & \textbf{50.0} & \textbf{70.1} \\
        \midrule
        \multirow{2}{*}{%
            \rotatebox[origin=c]{90}{\shortstack{ViT-\\B/14}}}
        & CRISP (w/o Grammar)
        & \underline{94.8} & \underline{82.5} & \underline{85.9}
        & \underline{63.9} & \underline{57.8} & \underline{77.0} \\
        & CRISP (full)
        & \textbf{97.0} & \textbf{84.4} & \textbf{88.0}
        & \textbf{67.0} & \textbf{58.5} & \textbf{79.0} \\
        \bottomrule
    \end{tabular}%
    }
\end{minipage}%
\hfill
\begin{minipage}[t]{0.45\textwidth}
    \centering
    \caption{Performance of predicate sets.}
    \label{tab:grammar_ablation_and_compaction}
    \vspace{-0.7em}
    \footnotesize
    \setlength{\tabcolsep}{2.5pt}
    \renewcommand{\arraystretch}{1.05}
    \resizebox{\linewidth}{!}{%
    \begin{tabular}{@{}lccccc@{}}
        \toprule
         & Art & Cartoon & Paint & Photo & Avg. \\
        \midrule
        No Rel.          & 66.3 & 62.0 & 45.8 & 68.6 & 60.7 \\
        Binary           & 69.6 & 68.7 & 47.9 & 75.3 & 65.4 \\
        Binary + Ternary & 71.7 & 71.5 & 50.5 & 76.8 & 67.6 \\
        All              & 72.4 & 72.3 & 51.9 & 79.6 & 69.0 \\
        \bottomrule
    \end{tabular}%
    }
\end{minipage}
\vspace{-0.8em}
\end{table*}

\section{Conclusion}\label{sec:conclusion}
We presented \method{}, an end-to-end domain generalization framework that combines learned visual primitives with differentiable spatial and appearance predicates. Class-specific sparse compositions of these predicates provide structural evidence, complemented by an appearance voting head operating on the same primitive regions.
Across five DomainBed datasets, \method{} achieves mean accuracies of 70.1\% with ResNet-50 and 79.0\% with ViT-B/14, improving over ERM++ by 1.2 and 0.4 percentage points, respectively. On CUB-DG with ResNet-50, it achieves 69.0\% mean accuracy, exceeding ERM++ by 7.9 percentage points. Predicate ablations show that binary and ternary spatial relations improve over primitive-only classification, while shared unary appearance predicates provide further gains. These findings support combining explicit relational composition with local appearance evidence for domain generalization.

\bibliographystyle{plainnat}
\bibliography{main}

\newpage
\appendix
\section*{Appendix}

\section{Dataset and Implementation details}\label{app:datasets}
The DomainBed consists of 5 datasets: \textsc{PACS}~\citep{li2017deeper} (4 domains, 7 classes, and 9,991 images), \textsc{VLCS}~\citep{fang2013unbiased} (4 domains, 5 classes, and 10,729 images), OfficeHome~\citep{venkateswara2017deep} (4 domains, 65 classes, and 15,588 images), TerraIncognita~\citep{beery2018recognition} (4 domains, 10 classes, and 24, 788 images), and DomainNet~\citep{peng2019moment} (6 domains, 345 classes, and 586,575 images). The CUB-DG dataset comprises 4 domains, with 47,152 images across 200 classes of North American bird species. All performance scores are evaluated by \textit{leave-one-out} cross-validation, where averaging all cases that use a single domain as the target (test) domain and the others as the source (training) domains. We hold out 20\% of the source domain data for validation and select the best validation model for testing on the held-out target domain. 
Since we use ResNet-50~\citep{he2016deep} backbone pretrained on ImageNet~\citep{deng2009imagenet} as backbone, we benchmark against prior methods that utilize the same backbone architecture for fair comparison, which are: DANN~\citep{Ganin2016DANN}, CDANN~\citep{Long2018CDAN}, GroupDRO~\citep{Sagawa2020GroupDRO}, MixUp~\citep{zhang2018mixup}, SagNet~\citep{nam2021sagnet}, CORAL~\citep{Sun2016DeepCORAL}, ERM++~\citep{Teterwak2025ERMPlusPlus}, MIRO~\citep{cha2022miro}, and SWAD~\citep{cha2021swad}. We also include GVRT~\citep{Min2022GVRT} as the current benchmark due to the remarkable SoTA performance on the CUB-DG dataset.

\noindent\textbf{Implementation details.}
The model is optimized using the Adam~\citep{kingma2014adam} optimizer with a default learning rate of $0.001$. We set the number of primitives to $K=16$ for all domains in the DomainBed and CUB-DG settings, while keeping the same batch size $N=32$ to be consistent with prior methods. The number of target angles $n$ and turning angles $\ell$ of ternary predicates are set to 3 and 1, respectively. 
Following prior DG work, we apply feature-level style mixing~\citep{zhou2021domain} during training to reduce sensitivity to domain-specific texture statistics. 
We achieve these hyperparameters via grid search, which is reproducible via the source code.
All of the experiments were conducted on a single RunPod cloud instance equipped with 9 vCPUs, 50 GB system RAM, and 60 GB of storage, and an NVIDIA RTX 4000 Ada Generation GPU with 20 GB GDDR6 VRAM. For the training time, it takes 0.8 seconds per step, resulting in approximately 440s per epoch on the CUB-DG dataset. 

\section{Regularization and Domain Adversarial Loss}\label{app:regularizers}

This section provides the full definition of the regularization and the adversarial term used in
Eq.~\ref{eq:learning_objective}. The full training objective is
\[
\mathcal{L}
=
\mathcal{L}_{\mathrm{CE}}(s(X),y) + \mathcal{L}_\text{reg} + \mathcal{L}_\text{DANN}
\]
where 
\[
\mathcal{L}_\text{reg} = 
\lambda_{\mathrm{sparse}}\,\overline{\|\Lambda\|}_1
+
\lambda_{\mathrm{bn}}\,\mathcal{L}_{\mathrm{bn}}
+
\lambda_{\mathrm{ang}}\,\mathcal{L}_{\mathrm{ang}},
\]
where $\mathcal{L}_{\mathrm{CE}}$ is the standard cross-entropy loss over structural
class scores, and the remaining terms are described below.
The predicate parameters receive no direct supervision
beyond $\mathcal{L}_{\mathrm{CE}}$; they are optimized entirely through the
gradient path from class scores through grammar weights to relation scores.

\paragraph{Sparsity regularization.}
The class-weight matrix $\Lambda\in\mathbb{R}^{|\mathcal{C}|\times M}$
holds class-specific log-weights over the $M$ enumerated predicate applications.
We apply a mean $\ell_1$ penalty,
\[
\overline{\|\Lambda\|}_1
=
\frac{1}{|\mathcal{C}|\,M}
\sum_{c=1}^{|\mathcal{C}|}\sum_{m=1}^{M}|\Lambda_{c,m}|,
\]
which complements sparsemax normalization by penalizing log-weight magnitude
before the activation, encouraging each class to rely on a compact subset of
predicate applications.

\paragraph{Bottleneck regularization.}
To encourage the concept bottleneck to learn spatially meaningful and non-redundant primitive detectors, we regularize the normalized primitive heatmaps $\{\widetilde{H}_{k}\}$.
We define
\[
\mathcal{L}_{\mathrm{bn}}
=
\mathcal{L}_{\mathrm{div}}
+
\lambda_{\mathrm{conc}}\mathcal{L}_{\mathrm{conc}},
\]
where $\mathcal{L}_{\mathrm{div}}$ discourages different primitives from attending to the same spatial regions, and $\mathcal{L}_{\mathrm{conc}}$ encourages each primitive heatmap to be spatially concentrated.
Concretely, flattening each normalized heatmap $\widetilde{H}_{k}$ into a vector in $\Delta^{H_FW_F-1}$, we compute
\[
\mathcal{L}_{\mathrm{div}}
=
\frac{1}{K(K-1)}
\sum_{k\neq k'}
\frac{
\widetilde{H}_{k}^{\top}\widetilde{H}_{k'}
}{
\|\widetilde{H}_{k}\|_2\,\|\widetilde{H}_{k'}\|_2
},
\quad
\mathcal{L}_{\mathrm{conc}}
=
-\frac{1}{K}\sum_{k=1}^{K}
\sum_{h,w}
\widetilde{H}_{k,h,w}\log (\widetilde{H}_{k,h,w} + \epsilon).
\]
Where $\epsilon = 0.01$ is the normalization term.
This bottleneck regularizer is used in every training run and helps reduce primitive collapse.

\paragraph{Domain Adversarial Features regularization.} For each pair of $(X, d)$, denote the input image with the corresponding source domain label; the goal of this regularization term is to make the extracted features $F$ from the backbone $f_{\theta}$ distributionally indistinguishable. Thus, following the work of~\citep{Ganin2016DANN}, we employ a Gradient Reversal Layer (GRL) with a domain classifier $f_{Dom}$. In practice, $f_{Dom}$ is implemented as a Linear classifier that predicts which domain the features come from:
$$
f_{Dom} (F) = \hat{d} \in \mathbb{R}^{|\mathcal{S}|}
$$
where $|\mathcal{S}|$ is the total number of source domains. To achieve domain confusion, we apply a standard cross-entropy loss $\mathcal{L}_\text{DANN} = \mathcal{L}_\text{CE}(\hat{d}, d)$ to the predictions of $f_{Dom}$.
During training, the GRL acts as an identity transformation in the forward pass. However, during the backward pass, it multiplies the gradient from $\mathcal{L}_\text{DANN}$ by a negative scalar $-\lambda$. This forces the domain classifier $f_{Dom}$ to minimize the classification error while simultaneously driving the backbone $f_{\theta}$ to maximize it, ultimately resulting in domain-invariant feature representations.

\section{Per-Dataset results}\label{app:per-domain}

\begin{figure}
    \centering
    \includegraphics[width=1\textwidth]{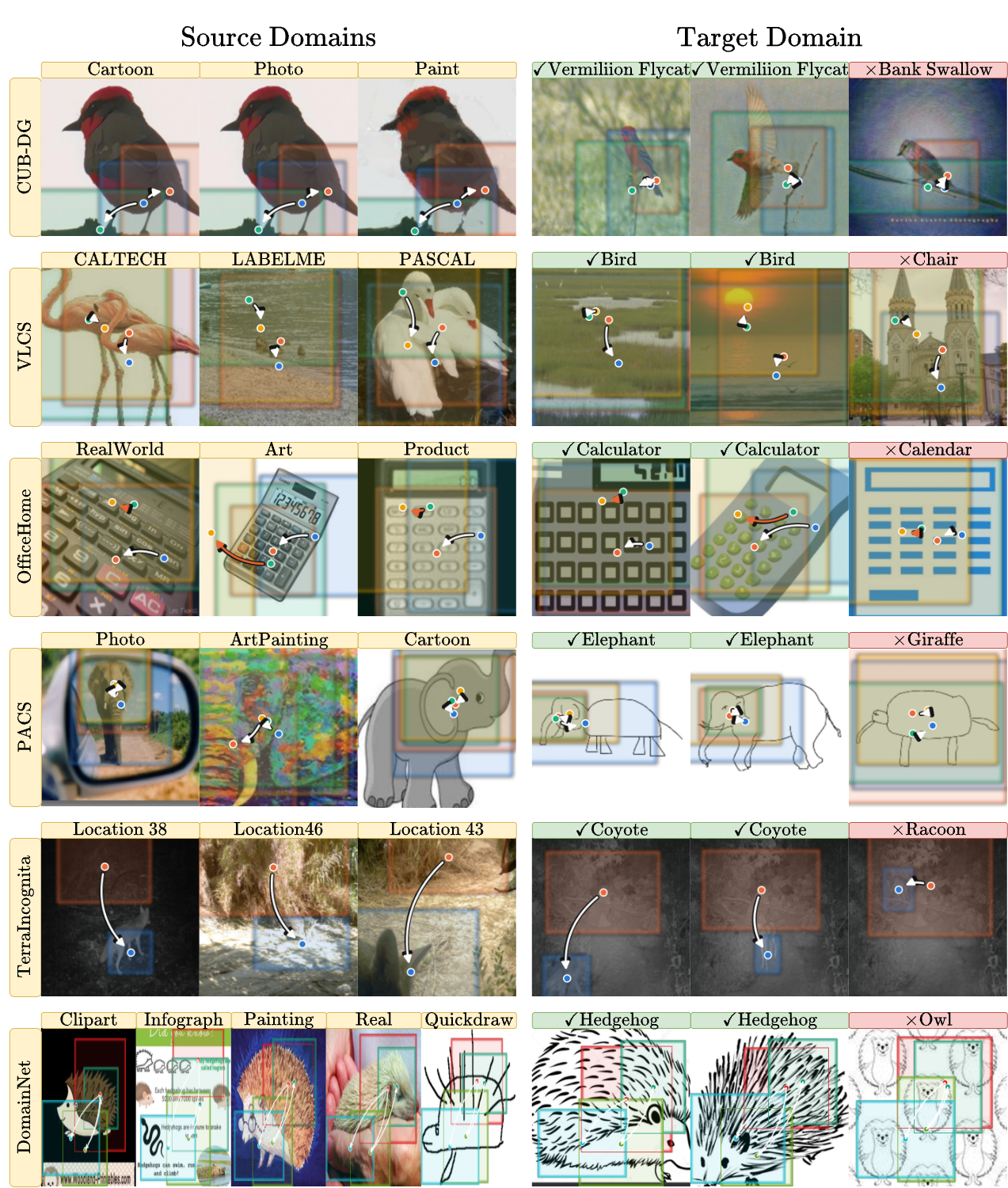}
    \caption{The visualization of samples from all 6 datasets: CUB-DG, VLCS, PACS, OfficeHome, TerraIncognita, and DomainNet. For each dataset, we visualize a class with the corresponding source samples, as well as two correctly and one incorrectly classified samples from the target domain. Furthermore, each image visualizes the most prominent production, with its primitives and relation, denoted with different centroids and bounding boxes.}
    \label{fig:all_domain_vis}
\end{figure}

\paragraph{Results on PACS}
Table~\ref{tab:pacs} reports per-domain accuracy on PACS. With ResNet-50, \method{} achieves the highest accuracy among the listed methods on Art painting and Photo, while its Sketch accuracy falls sharply to 7.1 percentage points compared with ERM++, resulting in a lower average of 88.4\% versus 89.8\%. With ViT-B/14, \method{} reaches 97.0\% average accuracy, slightly exceeding ERM++ by 0.2 percentage points. These results indicate that improvements are domain-dependent, particularly when transferring to sketches with the ResNet-50 backbone. As illustrated in Fig.~\ref{fig:all_domain_vis}, the correctly classified elephant sketches retain identifiable head and trunk regions whose spatial arrangement resembles that highlighted in the source examples despite substantial appearance changes. In the failed example predicted as a giraffe, the highlighted regions provide a less distinct head and trunk structure, suggesting that ambiguous primitive localization can weaken the relational evidence for classification.

\begin{table*}[t]
\centering
\small
\caption{Accuracy on PACS. Best in \textbf{bold}; second-best \underline{underlined}, within each backbone.}
\label{tab:pacs}
\setlength{\tabcolsep}{4.0pt}
\begin{tabular}{lccccc}
\toprule
Method & Art painting & Cartoon & Photo & Sketch & Avg. \\
\midrule
\multicolumn{6}{c}{\textbf{ResNet-50 Backbone}} \\
\midrule
ERM
& 84.7 & 80.8 & 97.2 & 79.3 & 84.2 \\
ERM + SWAD
& 89.3 & 83.4 & 97.3 & 82.5 & 88.1 \\
DIWA
& \underline{90.6} & 83.4 & 98.2 & \underline{83.8}
& \underline{89.0} \\
ERM + MIRO + SWAD
& -- & -- & -- & -- & 88.4 \\
ERM++
& \underline{90.6} & \underline{83.7} & 98.1
& \textbf{86.6} & \textbf{89.8} \\
ERM++ + MIRO
& 90.2 & \textbf{83.8} & \underline{98.6} & 82.4 & 88.8 \\
\midrule
\method{} (Ours)
& \textbf{91.9} & 83.2 & \textbf{98.8} & 79.5 & 88.4 \\
\midrule
\multicolumn{6}{c}{\textbf{ViT-B/14 Backbone}} \\
\midrule
ERM++
& -- & -- & -- & -- & \underline{96.8} \\
\midrule
\method{} (Ours)
& \textbf{99.1} & \textbf{96.8} & \textbf{100.0}
& \textbf{92.1} & \textbf{97.0} \\
\bottomrule
\end{tabular}
\end{table*}

\paragraph{Results on VLCS}
As illustrated in Table~\ref{tab:vlcs}, \method{} with ResNet-50 achieves the highest average accuracy of 80.4\%, outperforming ERM++ by 2.4 percentage points and the strongest listed baseline, ERM + MIRO + SWAD, by 0.8 points in the VLCS dataset. Relative to ERM++, the largest improvements occur on LabelMe and SUN09, with gains of 4.0 and 3.7 points, respectively. With ViT-B/14, \method{} improves average accuracy from 82.4\% to 84.4\%, showing gains over ERM++ with both backbones. In Fig.~\ref{fig:all_domain_vis}, the successful target examples show that the highlighted regions and their relations can support bird recognition even when the birds occupy a small portion of the scene. In the failed example predicted as a chair, the highlighted regions lie largely on architecture and surrounding scenery, suggesting that background structure can introduce misleading relational cues.

\begin{table*}[t]
\centering
\small
\caption{Accuracy on VLCS. Best in \textbf{bold}; second-best \underline{underlined}, within each backbone.}
\label{tab:vlcs}
\setlength{\tabcolsep}{4.0pt}
\begin{tabular}{lccccc}
\toprule
Method & Caltech101 & LabelMe & SUN09 & VOC2007 & Avg. \\
\midrule
\multicolumn{6}{c}{\textbf{ResNet-50 Backbone}} \\
\midrule
ERM
& 97.7 & \underline{64.3} & 73.4 & 74.6 & 77.3 \\
ERM + SWAD
& 98.8 & 63.3 & \textbf{75.3} & \underline{79.2} & 79.1 \\
DIWA
& 98.9 & 62.4 & \underline{73.9} & 78.9 & 78.6 \\
ERM + MIRO + SWAD
& -- & -- & -- & -- & \underline{79.6} \\
ERM++
& 98.7 & 63.2 & 71.6 & 78.7 & 78.0 \\
ERM++ + MIRO
& \underline{99.0} & 62.4 & 71.8 & 78.3 & 77.9 \\
\midrule
\method{} (Ours)
& \textbf{99.7} & \textbf{67.2} & \textbf{75.3}
& \textbf{79.4} & \textbf{80.4} \\
\midrule
\multicolumn{6}{c}{\textbf{ViT-B/14 Backbone}} \\
\midrule
ERM++
& -- & -- & -- & -- & \underline{82.4} \\
\midrule
\method{} (Ours)
& \textbf{99.6} & \textbf{69.6} & \textbf{81.6}
& \textbf{86.8} & \textbf{84.4} \\
\bottomrule
\end{tabular}
\end{table*}

\paragraph{Results on OfficeHome}
Table~\ref{tab:officehome} compares per-domain top-1 accuracy on OfficeHome. With ResNet-50, \method{} achieves 73.4\% average accuracy, exceeding ERM, ERM + SWAD, DIWA, and ERM + MIRO + SWAD, but remaining below ERM++ and ERM++ + MIRO. Although it improves Art accuracy over ERM++ by 1.0 percentage point, its Clipart accuracy decreases by 4.5 points, contributing to an overall reduction of 1.3 points. With ViT-B/14, \method{} achieves the highest reported average of 88.0\%, outperforming ERM++ by 1.3 points. The different outcomes across backbones suggest that the effectiveness of the primitive-based representation depends partly on the underlying visual features. As shown in Fig.~\ref{fig:all_domain_vis}, the correctly classified calculator examples preserve recognizable keypad regions and their spatial arrangement despite changes in rendering and viewpoint. The failed example predicted as a calendar also contains a regular grid of rectangular elements, illustrating how similar local primitives and layouts can create ambiguity between semantically different categories.

\begin{table*}[t]
\centering
\small
\caption{Accuracy on OfficeHome. Best in \textbf{bold}; second-best \underline{underlined}, within each backbone.}
\label{tab:officehome}
\setlength{\tabcolsep}{4.0pt}
\begin{tabular}{lccccc}
\toprule
Method & Art & Clipart & Product & Real & Avg. \\
\midrule
\multicolumn{6}{c}{\textbf{ResNet-50 Backbone}} \\
\midrule
ERM
& 63.1 & 51.9 & 77.2 & 78.1 & 67.6 \\
ERM + SWAD
& 66.1 & 57.7 & 78.4 & 80.2 & 70.6 \\
DIWA
& 69.2 & 59.0 & 81.7 & 82.2 & 73.0 \\
ERM + MIRO + SWAD
& -- & -- & -- & -- & 72.4 \\
ERM++
& 70.7 & \textbf{62.2} & \underline{81.8}
& \underline{84.0} & \underline{74.7} \\
ERM++ + MIRO
& \textbf{74.0} & \underline{61.5} & \textbf{83.8}
& \textbf{85.7} & \textbf{76.3} \\
\midrule
\method{} (Ours)
& \underline{71.7} & 57.7 & 81.1 & 83.0 & 73.4 \\
\midrule
\multicolumn{6}{c}{\textbf{ViT-B/14 Backbone}} \\
\midrule
ERM++
& -- & -- & -- & -- & \underline{86.7} \\
\midrule
\method{} (Ours)
& \textbf{88.0} & \textbf{79.6} & \textbf{91.4}
& \textbf{92.9} & \textbf{88.0} \\
\bottomrule
\end{tabular}
\end{table*}

\paragraph{Results on Terra Incognita}
\textbf{Terra Incognita.}
Table~\ref{tab:terra} reports accuracy across the four held-out locations of Terra Incognita. With ResNet-50, \method{} achieves the highest average accuracy of 58.5\%, improving over ERM++ by 7.3 percentage points and ERM++ + MIRO by 5.1 points. It obtains the best results at Locations 100, 43, and 46, with the largest improvement over ERM++ occurring at Location 100: 70.7\% versus 48.3\%. Performance at Location 38 remains below ERM++, indicating that the gains are not uniform across environments. With ViT-B/14, \method{} achieves 67.0\%, slightly below the 67.4\% obtained by ERM++.
In Fig.~\ref{fig:all_domain_vis}, the correctly classified coyote examples retain a spatial arrangement between a broad scene region and a localized animal region resembling that highlighted in the source images, despite changes in illumination and background. In the failed example predicted as a raccoon, the highlighted regions shift toward a small background area, suggesting that inaccurate localization can replace useful animal-related evidence with scene-specific cues.

\begin{table*}[t]
\centering
\small
\caption{Accuracy on Terra Incognita. Best in \textbf{bold}; second-best \underline{underlined}, within each backbone.}
\label{tab:terra}
\setlength{\tabcolsep}{4.0pt}
\begin{tabular}{lccccc}
\toprule
Method & Loc.~100 & Loc.~38 & Loc.~43 & Loc.~46 & Avg. \\
\midrule
\multicolumn{6}{c}{\textbf{ResNet-50 Backbone}} \\
\midrule
ERM
& 54.3 & 42.5 & 55.6 & 38.8 & 47.8 \\
ERM + SWAD
& 55.4 & 44.9 & 59.7 & 39.9 & 50.0 \\
DIWA
& 57.2 & \underline{50.1} & 60.3 & 39.8 & 51.9 \\
ERM + MIRO + SWAD
& -- & -- & -- & -- & 52.9 \\
ERM++
& 48.3 & \textbf{50.7} & \underline{61.8}
& \underline{43.9} & 51.2 \\
ERM++ + MIRO
& \underline{60.8} & 48.8 & 61.1 & 42.7 & \underline{53.4} \\
\midrule
\method{} (Ours)
& \textbf{70.7} & 49.2 & \textbf{64.7}
& \textbf{49.2} & \textbf{58.5} \\
\midrule
\multicolumn{6}{c}{\textbf{ViT-B/14 Backbone}} \\
\midrule
ERM++
& -- & -- & -- & -- & \textbf{67.4} \\
\midrule
\method{} (Ours)
& \textbf{77.1} & \textbf{56.3} & \textbf{68.8}
& \textbf{65.9} & \underline{67.0} \\
\bottomrule
\end{tabular}
\end{table*}

\paragraph{Results on DomainNet}
Table~\ref{tab:domainnet} presents per-domain top-1 accuracy on DomainNet. With ResNet-50, \method{} obtains the best results on Painting and Infograph, improving over ERM++ by 1.5 and 2.0 percentage points, respectively. However, these gains are offset by lower accuracy on Clipart, Quickdraw, and Sketch, resulting in an average of 50.0\%, compared with 50.8\% for ERM++. With ViT-B/14, \method{} reaches 58.5\%, also below the 59.8\% achieved by ERM++. Thus, although the method benefits some visual domains, the results do not demonstrate a consistent advantage across the diverse representations in DomainNet. As illustrated in Fig.~\ref{fig:all_domain_vis}, the correctly classified hedgehog sketches retain recognizable face and spiny-body regions whose relative arrangement remains visible across the source styles. In the failed example predicted as an owl, the highlighted regions span multiple owl drawings and repeated textures, suggesting that fragmented localization and superficially similar patterns can produce misleading primitive relations.

\begin{table*}[t]
\centering
\small
\caption{Accuracy on DomainNet. Best in \textbf{bold}; second-best \underline{underlined}, within each backbone.}
\label{tab:domainnet}
\setlength{\tabcolsep}{4.0pt}
\begin{tabular}{lccccccc}
\toprule
Method & Painting & Clipart & Info & Real & Quickdraw & Sketch & Avg. \\
\midrule
\multicolumn{8}{c}{\textbf{ResNet-50 Backbone}} \\
\midrule
ERM
& 50.1 & 63.0 & 21.2 & 63.7 & 13.9 & 52.9 & 44.1 \\
ERM + SWAD
& 53.5 & 66.0 & 22.4 & 65.8 & 16.1 & 55.5 & 46.6 \\
DIWA
& 55.4 & 66.2 & 23.3 & 68.7 & \underline{16.5} & 56.0 & 47.7 \\
ERM + MIRO + SWAD
& -- & -- & -- & -- & -- & -- & 47.0 \\
ERM++
& 58.4 & \textbf{71.5} & 26.2 & \underline{70.7}
& \textbf{17.3} & \textbf{60.5} & \textbf{50.8} \\
ERM++ + MIRO
& \underline{58.5} & \underline{71.0} & \underline{26.5}
& \textbf{71.1} & 15.9 & \underline{59.5} & \underline{50.4} \\
\midrule
\method{} (Ours)
& \textbf{59.9} & 69.0 & \textbf{28.2}
& \underline{70.7} & 14.3 & 57.6 & 50.0 \\
\midrule
\multicolumn{8}{c}{\textbf{ViT-B/14 Backbone}} \\
\midrule
ERM++
& -- & -- & -- & -- & -- & -- & \textbf{59.8} \\
\midrule
\method{} (Ours)
& \textbf{69.6} & \textbf{76.8} & \textbf{39.0}
& \textbf{79.9} & \textbf{18.6} & \textbf{67.0} & \underline{58.5} \\
\bottomrule
\end{tabular}
\end{table*}

\paragraph{Results on CUB-DG}
Table~\ref{tab:cub_dg} reports per-domain top-1 accuracy on CUB-DG. With ResNet-50, \method{} achieves the highest average accuracy of 69.1\%, exceeding the strongest listed baseline, ERM++, by 8.0 percentage points. It obtains the best results on Photo, Cartoon, and Art, outperforming the strongest competing result in each domain by 5.0, 8.1, and 13.5 points, respectively. However, its Paint accuracy is 13.9 points below ERM++, showing that these gains do not extend to every visual style. With ViT-B/14, \method{} improves Cartoon and Art accuracy over ERM++ but performs worse on Photo and Paint, yielding a lower average of 80.1\% versus 81.2\%. As demonstrated in Fig.~\ref{fig:all_domain_vis}, the correctly classified target birds retain highlighted primitives around the leg and tail regions, providing spatial cues that remain recognizable across changes in style and pose. In the failed example predicted as a bank swallow, the primitive regions include substantial background and perch information, which may obscure the relevant anatomical relations and contribute to the incorrect classification.

\begin{table*}[t]
\centering
\small
\caption{Accuracy on CUB-DG. Best in \textbf{bold}; second-best \underline{underlined}, within each backbone.}
\label{tab:cub_dg}
\setlength{\tabcolsep}{4.0pt}
\begin{tabular}{lccccc}
\toprule
Method & Photo & Cartoon & Art & Paint & Avg. \\
\midrule
\multicolumn{6}{c}{\textbf{ResNet-50 Backbone}} \\
\midrule
DANN
& 67.5 & 57.0 & 42.8 & 30.6 & 49.5 \\
CDANN
& 65.3 & 55.2 & 43.2 & 30.5 & 48.6 \\
GroupDRO
& 60.9 & 54.8 & 36.5 & 27.0 & 44.8 \\
MixUp
& 67.1 & 55.9 & 51.1 & 27.2 & 50.3 \\
SagNet
& 67.4 & 60.7 & 44.0 & 34.2 & 51.6 \\
CORAL
& 72.2 & 63.5 & 50.3 & 35.8 & 55.5 \\
GVRT (PTE)
& \underline{74.6} & \underline{64.2} & 52.2 & 37.0 & 57.0 \\
MIRO
& 60.1 & 49.9 & \underline{58.9} & 34.9 & 51.0 \\
SWAD
& 64.0 & 54.7 & 52.1 & 37.9 & 52.2 \\
ERM++
& 60.3 & 61.6 & 56.8 & \textbf{65.8} & \underline{61.1} \\
\midrule
\method{} (Ours)
& \textbf{79.6} & \textbf{72.3} & \textbf{72.4}
& \underline{51.9} & \textbf{69.1} \\
\midrule
\multicolumn{6}{c}{\textbf{ViT-B/14 Backbone}} \\
\midrule
ERM++
& \textbf{93.0} & \underline{82.4} & \underline{83.7}
& \textbf{65.7} & \textbf{81.2} \\
\midrule
\method{} (Ours)
& \underline{89.0} & \textbf{85.0} & \textbf{84.1}
& \underline{62.3} & \underline{80.1} \\
\bottomrule
\end{tabular}
\end{table*}

\newpage
\clearpage 

\end{document}